\documentclass{article}

\usepackage[a4paper, margin=.8752in]{geometry}

\usepackage[utf8]{inputenc} %
\usepackage[T1]{fontenc}    %
\usepackage{hyperref}       %
\usepackage{url}            %
\usepackage{booktabs}       %
\usepackage{amsfonts}       %
\usepackage{amsmath}
\usepackage{nicefrac}       %
\usepackage{microtype}      %
\usepackage{epsfig}
\usepackage{graphicx}
\usepackage{caption}
\usepackage{subcaption}
\usepackage{float}
\usepackage{multirow, eucal}
\usepackage{color}
\newcommand{\etal}{\textit{et al}.}
\newcommand{\eg}{\textit{e.g.}}
\newcommand{\ie}{\textit{i.e.}}

\def\bx{ {\bf{\widetilde{x}}} }
\def\bs{ {\bf{\widetilde{s}}} }

\usepackage[vlined,ruled,linesnumbered]{algorithm2e}

\title{Training Compact Neural Networks with Binary Weights and Low Precision Activations}

\author{
 Bohan Zhuang, Chunhua Shen, Ian Reid\\
 The University of Adelaide, Australia 
}
\date{}

\begin{document}

\maketitle

\begin{abstract}
In this paper, we propose to train a network with binary weights and low-bitwidth activations, designed especially for mobile devices with limited power consumption. Most previous works on quantizing CNNs uncritically assume the same architecture, though with reduced precision. However, we take the view that for best performance it is possible (and even likely) that a different architecture may be better suited to dealing with low precision weights and activations.
Specifically, we propose a ``network expansion'' strategy in which we aggregate a set of homogeneous low-precision branches to implicitly reconstruct the full-precision intermediate feature maps. Moreover, we also propose a group-wise feature approximation strategy which is very flexible and highly accurate. Experiments on ImageNet classification tasks demonstrate the superior performance of the proposed model, named Group-Net, over various popular architectures. In particular, with binary weights and activations, we outperform the previous best binary neural network in terms of accuracy as well as saving more than 5 times computational complexity on ImageNet with ResNet-18 and ResNet-50. 
\end{abstract}

\tableofcontents

\clearpage

\section{Introduction}

Designing deeper and wider convolutional neural networks has led to significant breakthroughs in many machine learning tasks, such as image classification~\cite{krizhevsky2012imagenet, he2016deep}, object detection~\cite{redmon2016you, ren2015faster} and object segmentation~\cite{long2015fully}. However, accuracy is roughly directly proportional to log(FLOPs)~\cite{canziani2016analysis, he2018adc} and training accurate CNNs always requires billions of FLOPs. However, running real-time applications on mobile platforms requires low energy consumption and high accuracy. To solve this problem, many existing works~\cite{he2017channel, han2015learning, zhuang2018towards, jacob2017quantization, chollet2017xception, howard2017mobilenets} focus on network pruning, low-bit quantization and efficient architecture design. In this paper, we aim to design a highly efficient low-precision neural network architecture from the quantization perspective.

Binary neural networks are first proposed in~\cite{hubara2016binarized, rastegari2016xnor} to accelerate inference and save memory usage. To improve the balance between accuracy and complexity, several works~\cite{guo2017network, li2017performance, lin2017towards} propose to employ tensor expansion to approximate the filters or activations while still possessing the advantage of binary operations. In particular, Guo~\etal~\cite{guo2017network} recursively performs residual quantization on pretrained full-precision weights and does convolution on each binary weight base. Similarly, Li~\etal~\cite{li2017performance} propose to expand the input feature maps into binary bases in the same manner. And Lin~\etal~\cite{lin2017towards} further expand both weights and activations but use a simple linear approach.
However, the weights and activations approximation above are only based on minimizing the local reconstruction error rather than considering the final loss. As a result, the quantization error will be accumulated during propagation which results in apparent accuracy drop especially on large scale datasets (\eg, ImageNet). What's more, they have to solve the linear regression problem for each layer during feedforward propagation and may suffer from rank deficiency if the bases are too correlated. In contrast, the proposed model is directly learnt to optimize the final objective while still implicitly taking into considerate the feature reconstruction in intermediate layers.

Interestingly, we are also motivated by the energy-efficient architecture design approaches~\cite{iandola2016squeezenet, howard2017mobilenets, zhang2017shufflenet}. The objective of all these approaches is to replace the traditional expensive convolution with computational efficient convolutional operations (\ie, depthwise seperable convolution, $1 \times 1$ convolution). Nevertheless, we propose to design extremely low-precision network architectures for dedicated hardware from the quantization view. Most previous quantization works focus on directly quantizing the full-precision architecture. At this point in time we do not yet learn the architecture, but we do begin to explore alternative architectures that we show are better suited to dealing with low precision weights and activations.
Specifically, we partition the full-precision model into groups and we decompose each group into a set of low-precision bases while still preserving the original network property. That is, with a bit more complexity and a little more memory space than directly quantizing the model, we can obtain near lossless quantization on the ImageNet dataset. What's more, the group-wise decomposition strategy (Fig.~\ref{fig:groupwise}) enjoys highly accurate, fast convergence and flexible properties which can be applied to any network structure in the literature (\eg, VGGNet, ResNet). 

We evaluate our models on the challenging CIFAR-100 and ImageNet datasets based on various architectures including AlexNet, ResNet-18 and ResNet-50. Extensive experiments show the effectiveness of the proposed method and the better performance over all the previous state-of-art quantization approaches. We expect that Group-Net will also generalize well on other recognition tasks.

\section{Related Work}
\vspace{-2mm}

\noindent\textbf{Network quantization:} The recent increasing demand for implementing fixed point deep neural networks on embedded devices motivates the study of network quantization. Several works have been proposed to quantize only parameters for highly accurate compression~\cite{li2016ternary, zhu2016trained, zhou2017incremental, courbariaux2015binaryconnect}. For example, Courbariaux \etal~\cite{courbariaux2015binaryconnect} propose to binarize the weights to replace multiply-accumulate operations by simple accumulations. Zhou \etal~\cite{zhou2017incremental} propose three interdependent operations, namely weight partition, group-wise quantization and re-training, to achieve lossless weights quantization. Further quantizing activations have also been extensively explored in literature~\cite{Cai_2017_CVPR, zhou2016dorefa, hubara2016binarized, rastegari2016xnor, lin2017towards, zhuang2018towards}. BNNs~\cite{hubara2016binarized} and XNOR-Net~\cite{rastegari2016xnor} propose to contrain both weights and activations to binary values (\ie, +1 and -1), where the multiply-accumulations can be replaced by the bitwise operations. To make a trade-off between accuracy and complexity, ~\cite{zhuang2018towards, zhou2016dorefa, hubara2016quantized, faraonesyq} experiment with different combinations of bitwidth for weights and activations and achieve improved accuracy compared to binary neural networks. 

\noindent\textbf{Efficient architecture design:} There has been a rising interest in designing efficient architecture in the recent literature. 
Efficient model designs like GoogLeNet~\cite{szegedy2015going} and SqueezeNet~\cite{iandola2016squeezenet} propose to replace 3x3 convolutional kernels with 1x1 size to reduce the complexity while increasing the depth and accuracy. Additionally, separable convolutions are also proved to be effective in Inception approaches~\cite{szegedy2016rethinking, szegedy2017inception}. This idea is further generalized as depthwise separable convolutions by Xception~\cite{chollet2017xception}, MobileNet~\cite{howard2017mobilenets} and ShuffleNet~\cite{zhang2017shufflenet} to generate energy-efficient network structure.
Recently, neural architecture search~\cite{zoph2016neural, pham2018efficient, zoph2017learning, liu2017progressive, liu2017hierarchical} using reinforcement learning has been explored for automatic model design. In particular, ENAS~\cite{pham2018efficient} greatly reduces the GPU-hours while still preserving the performance.

\section{Method}
\vspace{-2mm}
The objective of this paper is to binarize the weights and quantize the activations to low precision. To aid with discriptions, we adopt a terminology for architecture as comprising layers, blocks and groups. A layer is a standard single parameterized layer in a network such as a dense or convolutional layer, except with binary weights. A block is a collection of layers in which the output of end layer is connected to the input of the next block (\eg, a residual block). A group is a collection of blocks. We aim to explore two different architecture changes which we call layer-wise and group-wise. These are illustrated in Fig.~\ref{fig:overview} (respectively in (b) and (c)) along with a baseline architecture in Fig.~\ref{fig:overview} (a) that simply adopts the same architecture as its ``parent'' but with binarized weights and quantized activations. In Sec.~\ref{sec:function}, we describe our quantization operation on weights and activations, respectively. Then we describe the layer-wise approach in Sec.~\ref{sec:layerwise}. We further extend it to the flexible group-wise structure in Sec.~\ref{sec:groupwise}.

\vspace{-2mm}
\subsection{Quantization function}\label{sec:function}
For a convolutional layer, we define the input ${\bf{x}} \in {\mathbb{R}^{{c_{in}} \times {w_{in}} \times {h_{in}}}}$, weight filter ${\bf{w}} \in {\mathbb{R}^{c \times w \times h}}$ and the output ${\bf{y}} \in {\mathbb{R}^{{c_{out}} \times {w_{out}} \times {h_{out}}}}$, respectively.

\noindent\textbf{Quantization of weights}: Following ~\cite{rastegari2016xnor}, we estimate the floating-point weight ${\bf{w}}$ by a binary weight filter ${\bf{b}}$ and a scaling factor $\alpha  \in {\mathbb{R}^ + }$ such that ${\bf{w}} \approx \alpha {\bf{b}}$, where ${\bf{b}}$ is the sign of ${\bf{w}}$ and $\alpha$ calculates the mean of absolute values of ${\bf{w}}$. In general, the quantization function is non-differentiable and we adopt the straight-through estimator~\cite{bengio2013estimating} (STE) to approximate the gradient calculation. 
Formally, the forward and backward process can be given as follows: 
\setlength{\abovedisplayskip}{1.5pt} 
\setlength{\belowdisplayskip}{1.5pt}
\begin{equation}
\begin{split}
&\mbox{Forward}:{\bf{b}} = \alpha \cdot {\rm{sign}}({\bf{w}}),\\
&\mbox{Backward}:\frac{{\partial \ell}}{{\partial {\bf{w}}}} = \frac{{\partial \ell}}{{\partial {\bf{b}}}}\cdot\frac{{\partial {\bf{b}}}}{{\partial {\bf{w}}}} \approx  \frac{{\partial \ell}}{{\partial {\bf{b}}}},\\
\end{split}
\end{equation}
where $\ell$ is the loss. In practice, we find such a binarization scheme is quite stable.

\noindent\textbf{Quantization of activations}: As the output of the RELU function is unbounded, the quantization after RELU requires a high dynamic range. It will cause large quantization errors especially when the bit-precision is low. To alleviate this problem, similar to~\cite{zhou2016dorefa, hubara2016quantized}, we use a clip function $h(y) = {\rm{clip}}(y,0,\beta )$ to limit the range of activation to $[0,\beta]$, where $\beta$ (not learned) is fixed during training. Then the truncated activation output ${\bf{y}}$ is linearly quantized to $k$-bits ($k > 1$) and we still use STE to estimate the gradient:
\setlength{\abovedisplayskip}{0.5pt} 
\setlength{\belowdisplayskip}{1.5pt}
\begin{equation}
\begin{split}
&\mbox{Forward}: {\bf{\widetilde{y}}} = {\rm{round}}({\bf{y}} \cdot \frac{{{2^k} - 1}}{\beta }) \cdot \frac{\beta }{{{2^k} - 1}}, \\
&\mbox{Backward}: \frac{{\partial \ell}}{{\partial {\bf{y}}}} = \frac{{\partial \ell}}{{\partial {\bf{\widetilde{y}}}}}. \\
\end{split}
\end{equation}
Note that when $k=1$, we follow the quantization scheme in XNOR-Net~\cite{rastegari2016xnor} by introducing scale factors for both weights and activations during binarization to preserve the accuracy. 

\vspace{-2mm}
\subsection{Layer-wise  feature reconstruction} \label{sec:layerwise}
\begin{figure*}[h]
	\centering
	\resizebox{0.9\linewidth}{!}
	{
		\begin{tabular}{c}
			\includegraphics{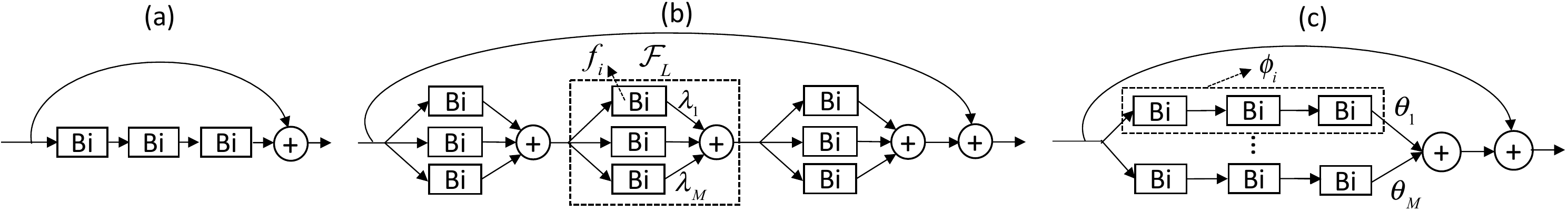}
		\end{tabular}
	}
	\caption{Overview of the baseline quantization method and two proposed approaches. We take one residual block for illustration. Bi is a binary weight layer. (a): Directly quantize the full-precision block. (b): Reconstruct full-precision activations layer-wise using a set of binary weight convolutional layers. All activations are quantized. (c): Approximate full-precision activations group-wise with a linear combination of binary weight bases ${\phi_i}$. All activations are quantized.}
	\label{fig:overview}
\end{figure*}

Fig.~\ref{fig:overview} (b) illustrates the layer-wise feature reconstruction for a single block. 
At each layer, we aim to reconstruct the full-precision output feature map given the input 3-D tensor $\bf{x}$ using a set of quantized homogeneous branches:
\begin{equation}\label{eq:1}
{{\mathcal{F}}_L}({\bf{x}}) = \sum\limits_{i = 1}^M {{\lambda _i}{f_i}({\bf{x}})}, 
\end{equation} 
where $f(\cdot)$ is the convolutional operation, $M$ is the number of branches and ${\lambda_i}$ is a scale factor. Note that when activations are quantized to $k>1$ bit, operation $f(\cdot)$ is just simple fixed-point accumulations similar to BinaryConnect~\cite{courbariaux2015binaryconnect}. When activations are also constrained to binary values (\ie, -1 or +1), $f(\cdot)$ is the bitwise operations: xnor and bitcount~\cite{zhou2016dorefa}. Note that the convolution with each binary filter can be computed in parallel. We explore both effects in Sec.~\ref{sec:bitwidth}.

All ${{f_i}}$'s in Eq.~\ref{eq:1} have the same convolution hyperparameters as the original floating-point counterpart. In this way, each low-precision branch gives a rough transformation and all the transformations are aggregated to approximate the original full precision output feature map. It can be expected that with the increase of $M$, we can get more accurate approximation with more complex transformations. A special case is when $M=1$, it corresponds to directly quantize the full-precision network.
  
However, this strategy has an apparent limitation. The quantization function in each branch introduces a certain amount of error. Furthermore, as the previous estimation error propagates into the current multi-branch convolutional layer, it will be enlarged by the quantization function and the final aggregation process. As a result, it may cause large quantization errors especially for deeper layers and bring large deviation for gradients during backpropagation. To solve this problem, we further propose a flexible group-wise approximation approach in Sec.~\ref{sec:groupwise}. 

\emph{Complexity}: We consider the binary convolution case here ($k = 1$) where operations are XNOR bit counts. One floating-point operation roughly equals to 64 binary operations within one clock cycle. We need to calculate $M$ binary convolutions and $M$ full-precision additions, thus the speed up ratio $\sigma$ can be calculated as:
\begin{equation}
\sigma  = \frac{{{c_{{\rm{in}}}}{c_{out}}wh{w_{in}}{h_{in}}}}{{\frac{1}{{64}}(M{c_{{\rm{in}}}}{c_{out}}wh{w_{in}}{h_{in}}) + M{c_{out}}{w_{out}}{h_{out}}}} = \frac{{64}}{M} \cdot \frac{{{c_{{\rm{in}}}}{c_{out}}wh{w_{in}}{h_{in}}}}{{{c_{{\rm{in}}}}{c_{out}}wh{w_{in}}{h_{in}} + 64{c_{out}}{w_{out}}{h_{out}}}}
\end{equation}
This equation is valid for the VGG-style architecture that repeatedly stacks layers with the same shape or a ResNet bottleneck block (except for the subsampling layers). And for small enough non-binary bitwidth of activations (\eg, $k=2, 4$), the complexity will relatively increase but fixed-point addition is still very efficient for digital chips. The choice of $k$ depends on the practical demands. 

\subsection{Group-wise feature approximation}\label{sec:groupwise}

\begin{figure*}[!htb]
	\centering
	\resizebox{0.985\linewidth}{!}
	{
		\begin{tabular}{c}
			\includegraphics{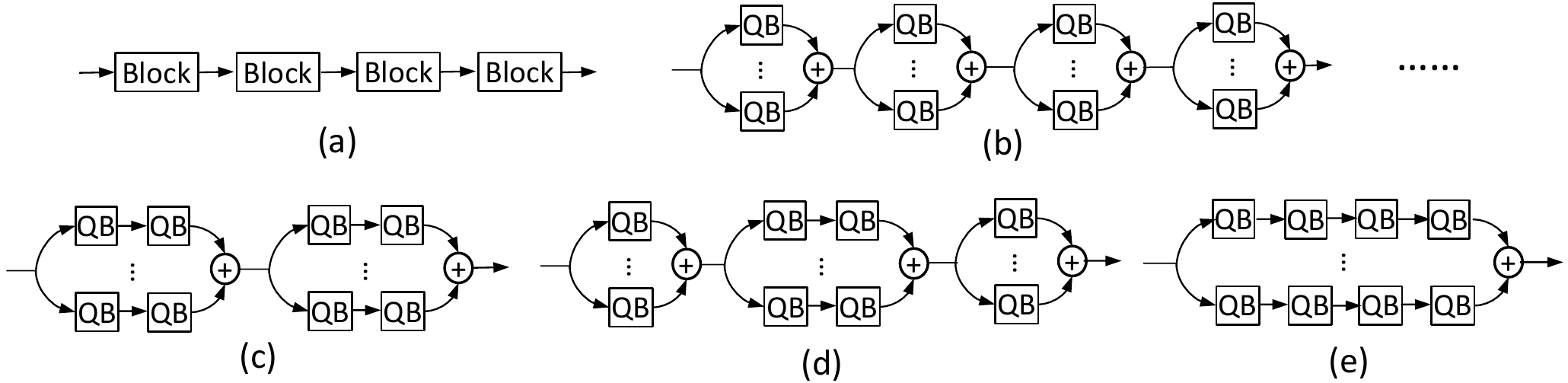}
		\end{tabular}
	}
	\caption{Illustration of several possible group-wise architectures. We assume the original full-precision network comprises four blocks as shown in (a). QB is the abbreviation of quantized block. We omit the skip connection for convenience. (b): Each group comprises one block in (a) and we approximate each floating-point block's output with a set of quantized blocks. (c): We partition two blocks as a group. (d): Approximate features using flexible block combination. (e): An extreme case. We ensemble a set of quantized networks to approximate the output before the classification layer.}
	\label{fig:groupwise}
\end{figure*}

In layer-wise approach, we approximate each layer separately. However, in this section we explore the idea of approximating an entire group. The motivation is as follows. As explained in Sec.~\ref{sec:layerwise}, the frequency of reconstructing the intermediate activations and suppressing the quantization error is a trade-off. Analogy to the extreme ``low-level'' layer-wise case, we also show the extreme ``high-level'' case in Fig.~\ref{fig:groupwise} (e) where we directly ensemble a set of low-precision networks. In the ensemble model, the output quantization error for each branch can be very large even though we only aggregate the outputs once. Clearly, these two extreme cases may not be the optimal and it motivates us to explore the ``mid-level'' cases. More specifically, we propose a group-wise approximation strategy which intends to approximate residual blocks or several consequent layers as a whole. In other words, each group consists of one or multiple residual blocks or even VGG-style plain structure by stacking layers without skip connections. But in this paper, we analyze on the residual structure for convenience. 

We first consider the simplest case where each group consists of only one block (\ie, the group comprises one block of Fig.~\ref{fig:groupwise} (a)). Then the above layer-wise approximation method can be easily extended to the group-wise structure. The most typical block structure is the bottleneck architecture and we can extend Eq.~\ref{eq:1} as:
\setlength{\abovedisplayskip}{1pt} 
\setlength{\belowdisplayskip}{1pt}
\begin{equation} \label{eq:2}
{{\mathcal{F}}_G}({\bf{x}}) = \sum\limits_{i = 1}^M {{\theta _i}{\phi _i}({\bf{x}})}  + {\bf{x}},
\end{equation}
where ${\phi(\cdot)}$ is a low-precision residual bottleneck~\cite{he2016deep} and ${\theta_i}$ is the scale factor. In Eq.~\ref{eq:2}, we use a linear combination of homogeous low-precision bases to approximate one group, where each base ${\phi _i}$ has one quantized block (QB in Fig.~\ref{fig:groupwise}). We illustrate such a group in Fig.~\ref{fig:overview} (c) and the framework which consists of these groups in Fig.~\ref{fig:groupwise} (b). In this way, we effectively keep the original residual structure in each base to preserve the network capacity. In addition, we keep a balance between supressing quantization error accumulation and feature reconstruction. Moreover, compared to layer-wise strategy, the number of parameters and complexity both decrease since there is no need to apply float-precision tensor aggregation within the group. Interestingly, the multi-branch group-wise design is parallelizable and hardware friendly, which can bring great speed-up during test-time inference. 

Furthermore, the group-wise approximation is flexible. We now analyze the case where each group may contain different number of blocks. Suppose we partition the network into $P$ groups and it follows a simple rule that the each group must include one or multiple complete residual building blocks.
Let $\{ {G_1},{G_2},...,{G_P}\}$ be the group indexes at which we approximate the output feature maps. For the $p$-th group, we consider the blocks $B \in \{ {B_{p - 1}} + 1,...,{B_p}\}$, where the index ${B_{p - 1}}=0$ if $p=1$. And we can extend Eq.~\ref{eq:2} into multiple blocks format:
\setlength{\abovedisplayskip}{1pt} 
\setlength{\belowdisplayskip}{1pt}
\begin{equation}\label{eq:3}
{\widetilde{{\cal F}_G}}({{\bf{x}}_{{B_{p - 1}} + 1}}) = \sum\limits_{i = 1}^M {{\theta _i}u_i^{{B_p}}(u_i^{{B_p} - 1}(...(u_i^{{B_{p - 1}} + 1}({{\bf{x}}_{{B_{p - 1}} + 1}}))...)} ),
\end{equation}
where $u(\bf{x}) = \phi (\bf{x}) + \bf{x}$ is the residual function~\cite{he2016deep}, except with binary weights and quantized activations. Based on ${\widetilde{{\mathcal{F}}}_G}$, we can efficiently construct a network by stacking these groups and each group may consist of one or multiple blocks. Different from Eq.~\ref{eq:2}, we further expose a new dimension on each base, which is the number of blocks. This greatly increases the flexibility of framework and the optimal structure can be effectively searched with reinforcement learning. 
We illustrate several possible connections in Fig.~\ref{fig:groupwise} and provide detailed discussions in Sec.~\ref{sec:discussion}.

\emph{Relation to ResNeXt~\cite{xie2017aggregated}}:
The homogeneous multi-branch architecture design shares some spirit of ResNeXt and enjoys the advantage of introducing a ``cardinality'' dimension. However, our objectives are totally different. ResNeXt aims to increase the capacity while maintaining the complexity. To achieve this, it first divides the input channels into groups and perform efficient group convolutions implementation. Then all the group outputs are aggregated to approximate the original feature map. In contrast, we first divide the network into groups and directly replicate the floating-point structure for each branch while both weights and activations are quantized. In this way, we can reconstruct the full-precision outputs via aggregating a set of low-precision transformations for complexity reduction in energy-efficient hardwares. Furthermore, our transformations are not restricted to only one block as in ResNeXt.

\emph{Relation to ShuffleNet~\cite{zhang2017shufflenet}}:
Based on ResNeXt, ShuffleNet proposes to increase the relation between groups by replacing the traditional group convolution with pointwise group convolution (GConv) and channel shuffle. We can draw an analogy between a convolution group and a branch $\phi_i$, where our each branch operates on all input channels. The last $1 \times 1$ GConv layer in the ShuffleNet unit applies a different scale on each output channel and then concatenate them together. However, output channels of $\phi_i$ share the same scale and we then simply add all branches' outputs as the final representation of the layer. The number of parameters of our group module increases but all the filters are binary and activations are quantized, which is still suitable for small storage and fast inference.

\emph{Relation to tensor expansion approaches~\cite{guo2017network, li2017performance, lin2017towards}}: 
In~\cite{guo2017network, lin2017towards}, binary weight bases are directly obtained from the full-precision weights without being learned. And they have to solve the linear regression problem for each layer during forward propagation. In contrast, we don't directly approximate the full-precision weights. Specifically, the binary weights are end-to-end optimized to minimize the final objective while still implicitly reconstructing the intermediate output feature maps. And in~\cite{li2017performance}, the input tensor for each layer is decomposed into binary residual tensors and convolved with shared binary weights.
However, it cannot guarantee the good approximation of the layer's output and introduces additional float-precision tensor additions at the beginning of each layer. 
In contrast to tensor expansion, we propose to approximate the full-precision network via ``network expansion". Futhermore, our network structure is quite flexible and highly accurate. 

\vspace{-2mm}
\subsection{Discussion} \label{sec:discussion}
The group-wise approximation approach can be efficiently integrated with the Neural Architecture Search (NAS) frameworks~\cite{zoph2016neural, pham2018efficient, zoph2017learning, liu2017progressive, liu2017hierarchical} to explore the optimal architecture. In our setting, the architecture hyperparameters that we need to generate are the number of groups and how we partition the blocks into these groups. We assume that the network has $Z$ blocks and we partition it into $P$ groups. Then the search space is ${2^{Z - 1}}$. For practical networks, $Z$ is usually not very large. So our search space is much smaller than those that need to generate each layer's hyperparameters (\eg, ${6^L} \times {2^{L(L - 1)/2}}$ for $L$ layers in ENAS~\cite{pham2018efficient}).
The proposed approach can also be combined with knowledge distillation strategy as in~\cite{zhuang2018towards, romero2015fitnets}. The basic idea is to train a target network alongside another guidance network. An additional regularizer is added to minimize the difference between student's and teacher's intermediate feature representations for higher accuracy. 
However, in this paper we only focus on designing efficient low-precision networks and leave these incremental tasks for future extension. We summarize the training algorithm in Sec. S2 in the supplementary material.

\section{Experiment}
\vspace{-2mm}
The proposed method is evaluated on CIFAR-100~\cite{krizhevsky2009learning} and ImageNet (ILSVRC2012) \cite{russakovsky2015imagenet} datasets. CIFAR-100 is an image classification benchmark containing images of size $32 \times 32$ in a training set of 50,000 and a test set of 10,000. ImageNet is a large-scale dataset which has $\sim$1.2M training images from 1K categories and 50K validation images. The evaluation metrics are top-1 and top-5 classification accuracy. Several representative networks are tested: AlexNet~\cite{krizhevsky2012imagenet} and ResNet~\cite{he2016deep}. Our implementation is based on Pytorch~\cite{paszke2017automatic}. 

To investigate the performance of the proposed methods, we analyze the effects of the number of bases, different group architectures and the difference between group-wise approximation and layer-wise approximation strategies. We define several methods for comparison as follows: 

\noindent\textbf{Layerwise:} It implements the layer-wise feature approximation strategy described in Sec.~\ref{sec:layerwise}.
\noindent\textbf{Group-Net v1:} We implement with the group-wise feature approximation strategy, where each base consists of one block. It corresponds to the approach described in Eq.~\ref{eq:2} and is illustrated in Fig.~\ref{fig:groupwise} (b).
\noindent\textbf{Group-Net v2:} Similar to Group-Net v1, the only difference is that each group base has two blocks. It is illustrated in Fig.~\ref{fig:groupwise} (c) and is explained in Eq.~\ref{eq:3}.\\
\noindent\textbf{Group-Net v3:} It is an extreme case where each base is a whole network, which can be treated as an ensemble of a set of low-precision networks. This case is shown in Fig.~\ref{fig:groupwise} (e). 
\vspace{-3mm}
\subsection{Implementation details}
As in \cite{rastegari2016xnor, Cai_2017_CVPR, zhou2016dorefa, zhuang2018towards}, we quantize the weights and activations of all layers except that the first and last layers have full precisions. In all ImageNet experiments, images are resized to $256 \times 256$, and a $224 \times 224$ ($227 \times 227$ for AlexNet) crop is randomly sampled from an image or its horizontal flip, with the per-pixel mean subtracted. We do not use any further data augmentation in our implementation. Batch Normalization is applied before each quantization layer as in \cite{Cai_2017_CVPR, zhou2016dorefa, zhuang2018towards}. We use a simple single-crop testing for standard evaluation. No bias term is utilized. We use Nesterov momentum SGD for optimization. 
For training all low-bitwidth networks, the mini-batch size and weight decay are set to 128 and 0.0001, respectively. The momentum ratio is 0.9. For training ResNet with non-binary activations, the learning rate starts at 0.05 and is divided by 10 when it gets saturated. We directly learn from scratch since we empirically observe that fine-tuning does not bring further benefits to the performance. But for training ResNet with binary activations, we decrease the learning rate to 0.001 to avoid frequent sign changes and we pretrain its full-precision counterpart for initialization to guarantee the accuracy. We remove the nonlinear function before the classification layer for ResNet based cases. For AlexNet, we directly learn from scratch and the initial learning rate is set to 0.01. Following~\cite{Cai_2017_CVPR, zhuang2018towards}, no dropout is used due to quantization itself can be treated as a regularization.

\vspace{-2mm}
\subsection{Evaluation on ImageNet} \label{sec:imagenet}

\setlength{\abovedisplayskip}{1pt} 
\setlength{\belowdisplayskip}{1pt}
\begin{table*}[h]
	\centering
	\scalebox{0.8}
	{
		\begin{tabular}{c| c | c c c c c c c c c}
			\hline
			\multicolumn{2}{c|}{model} &Full &BNN &XNOR &DOREFA &HWGQ & EL-Net &SYQ &ABC-Net &Ours \\\hline
			\hline
			\multirow{3}{*}{AlexNet}& Top-1 &57.2\% &27.9\%  &44.2\%  &46.4\%  &52.7\% &52.5\% &56.5\%  &-   &\bf{57.3}\%	\\ 
			&Top-5 &80.3\% &50.4\% &69.2\% &76.8\% &76.3\% &77.3\% &79.4\% &- &\bf{80.1}\% \\
			&Top-1 gap &- &29.3\% &13.0\% &10.8\% &4.5\% &4.7\% &0.7\% &- &\bf{-0.1}\%  \\\hline	 
			\multirow{3}{*}{ResNet-18}& Top-1 &69.7\% &42.2\% &51.2\% &60.9\% &59.6\% &-  &62.9\%  &65.0\% &\bf{67.6}\%\\ 
			&Top-5 &89.4\% &67.1\% &73.2\% &82.4\% &82.2\% &- &84.6\% &85.9\% &\bf{87.8}\% \\
			&Top-1 gap &- &27.5\% &18.5\% &8.8\% &10.1\% &- &6.8\% &4.7\% &\bf{2.1}\%\\\hline	 		
			\multirow{3}{*}{ResNet-50}& Top-1 &76.0\% &-  &-  &67.1\% &64.6\% &70.8\% &70.6\% &70.1\% &\bf{73.4}\%  \\ 
			&Top-5 &92.9\% &- &- &87.3\% &85.9\% &88.3\%  &89.6\% &89.7\% &\bf{90.8}\%\\
			&Top-1 gap &- &-  &- &8.9\% &11.4\% &5.2\% &5.4\% &5.9\% &\bf{2.6}\% \\\hline	 		         	 
			
	\end{tabular}}
	\caption{Comparison with the state-of-the-art low-precision methods using AlexNet,  ResNet-18 and ResNet-50 on ImageNet. Top-1 gap to the corresponding full precision networks is also reported. All the comparing results are directly cited from the original papers.}
	\label{tab:compare}
\end{table*}

In Table~\ref{tab:compare}, we compare our approach with the state-of-the-art quantization approaches BNN~\cite{hubara2016binarized}, XNOR-Net~\cite{rastegari2016xnor}, DOREFA-Net~\cite{zhou2016dorefa}, HWGQ-Net~\cite{Cai_2017_CVPR}, EL-Net~\cite{zhuang2018towards}, SYQ~\cite{faraonesyq} and ABC-Net~\cite{lin2017towards}, on ImageNet image classification task. 
For comparison, we consider the AlexNet, ResNet-18 and ResNet-50 in this section. 
\textbf{In all cases, our model uses 5 group bases with binary weights and 2-bit activations}. 
The results for HWGQ-Net is also based on binary weights and 2-bit activations. DOREFA-Net and EL-Net use 2-bit weights and 2-bit activations. SYQ employs binary weights and 8-bit activations.  ABC-Net is with 5 binary weight bases and 5 binary activation bases, which needs to calculate 25 times binary convolution and 6 times floating-point tensor accumulation in each layer. All the comparison results are directly cited from the corresponding papers (except DOREFA-Net is based on self-implementation). We also report the full-precision accuracy for all comparing models by our implementation. For ResNet, we use \emph{Group-Net v1} for comparison. For AlexNet, we treat two subsequent convolutional or dense layers as a group and employ the group-wise quantization strategy. From Table~\ref{tab:compare}, we can observe that the proposed method outperforms all the previous state-of-the-art by a large margin. It proves that the proposed approach can learn to approximate the full-precision network effectively. With binary weights and 2-bit activations, the proposed approach performs quite stably on all compared popular network structures. With more sophisticated quantization methods~\cite{hou2017loss, hou2018loss} and optimization~\cite{zhuang2018towards, zhu2016trained}, we expect that our performance may be further improved for more practical applications. 

\vspace{-2mm}
\subsection{Ablation study}
The core idea of Group-Net is the group-wise feature reconstruction strategy and we evaluate it comprehensively in this subsection on ImageNet dataset with ResNet-18 and ResNet-50.

\vspace{-2mm}
\subsubsection{Bitwidth impact} \label{sec:bitwidth}
\vspace{-2mm}
\begin{table*}[h]
	\centering
	\scalebox{0.85}
	{
		\begin{tabular}{c c c c c c c}
			\hline
		    Model &W&A&Top-1&Top-5&Top-1 gap&Top-5 gap\\
			 ResNet-18 Full-precision & 32  &32  &69.7\% &89.4\% &- &-  \\
			 ResNet-18 Group-Net v1 &  1  &1  &65.2\%  &85.6\%   &4.5\%  &3.8\%\\
			 ResNet-18 Group-Net v1 &1 &2 &67.6\% &87.8\% &2.1\% &1.6\% \\
		     ResNet-18 Group-Net v1 & 1 &4  &69.2\%  &88.5\%  &0.5\%  &0.9\%\\
		     ResNet-18 Group-Net v1 & 1 & 32 & 69.6\% & 89.1\% & 0.1\%  &0.3\% \\\hline	
		     \hline	
		     ResNet-18 Group-Net v2 &1 &4 &68.3\% &87.9\% &1.4\% &1.5\% \\
		     ResNet-18 Group-Net v3 &1 &4 &64.5\% &85.0\% &5.2\% &4.4\% \\
		     ResNet-18 Layerwise &1 &4 &60.1\% &82.2\% &9.6\% &7.2\%  \\\hline
		     \hline
		     ResNet-50 Full-precision &32 &32 &76.0\% &92.9\% &- &- \\
		     ResNet-50 Group-Net v1 &1  &1  &70.4\% &89.0\%  &5.6\%  &3.9\% \\
		     ResNet-50 Group-Net v1 &1 &2 &73.4\% &90.8\% &2.6\% &2.1\% \\
		     ResNet-50 Group-Net v1 &1 &4 &75.2\% &91.7\% &0.8\% &1.2\% \\\hline
	\end{tabular}}
	\caption{Validation accuracy of Group-Net on ImageNet with different choices of W and A for \emph{Group-Net v1}. `W' and `A' refer to the weight and activation bitwidth, respectively. Effect of different group divisions is also reported as \emph{Group-Net v2} and \emph{Group-Net v3}. The result of \emph{Layerwise} on ResNet-18 is also provided.}
	\label{tab:bitwidth}
\end{table*}
\vspace{-2mm}
This set of experiment is performed to assess the influence of activation precision for the final accuracy. We take the \emph{Group-Net v1} with ResNet-18 and ResNet-50 on ImageNet for analysis. We still use 5 group bases for experiment. The results are provided in Table~\ref{tab:bitwidth}. Results show that with binary weights and 4-bit or full-precision activations, we can achieve near lossless accuracy. For example, with binary weights and 4-bit activations, the top-1 accuracy drops are only \textbf{0.5\%} and \textbf{0.8\%} for ResNet-18 and ResNet-50, respectively. Interestingly, with binary activations where convolutional operations are all XNOR and bitcount, we can achieve comparable performance with ABC-Net, but that method has more than \boldmath$5 \times$ complexity compared with our group-wise design.

\vspace{-2mm}
\subsubsection{Effect of the number of bases}
\vspace{-2mm}
\begin{table*}[h]
	\centering
	\scalebox{0.85}
	{
		\begin{tabular}{c c c c c c c}
			\hline
			Model &Bases &Top-1&Top-5&Top-1 gap&Top-5 gap\\
			Full-precision &1  &69.7\% &89.4\% &- &-  \\
			Group-Net v1 &1  &60.1\%  &81.8\%   &9.6\% &7.6\%\\
			Group-Net v1 &3  &67.6\% &87.5\% &2.1\% &1.9\% \\
			Group-Net v1 &5 &69.2\%  &88.5\%  &\textbf{0.5}\% &\textbf{0.9}\%\\\hline	
	\end{tabular}}
	\caption{Validation accuracy of Group-Net on ImageNet with different group bases. All cases are based on the ResNet-18 network with binary weight and 4-bit activations.}
	\label{tab:bases}
\end{table*}
\vspace{-3mm}
We further explore the influence of the number of group bases to the final performance in Table~\ref{tab:bases} and Fig.~\ref{fig:convergence} (a). When the base is set to 1, it corresponds to directly quantize the original full-precision network and we observe apparent accuracy drop compared to its full-precision counterpart. With more bases employed, we can find the performance steadily increase. The reason can be attributed to the better approximation of the output feature maps, which is a trade-off between accuracy and complexity. It can be expected that with enough bases, the network should has the capacity to approximate the full-precision network precisely. With the multi-branch group-wise design, we can achieve high accuracy while still significantly reducing inference time and power consumption. Interestingly, each base can be implemented using small resource and the parallel structure is quite friendly to FPGA. 

\vspace{-2mm}
\subsubsection{Group space exploration}
We are also interested in exploring the influence of the number of blocks in each group base. We present the results in Table~\ref{tab:bitwidth} and Fig.~\ref{fig:convergence} (b). We observe that by approximating the output feature maps for each block results in the best performance on ResNet-18. It proves that by approximating appropriate intermediate layers, the classification accuracy will increase. However, this connection may not be the optimal. We expect to further boost the performance by integrating with the NAS approaches as discussed in Sec.~\ref{sec:discussion}.

\vspace{-1mm}
\subsubsection{Layer-wise vs. group-wise} \label{sec:vs}
\begin{figure*}[h]
	\centering
	\resizebox{0.85\linewidth}{!}
	{
		\begin{tabular}{c}
			\includegraphics{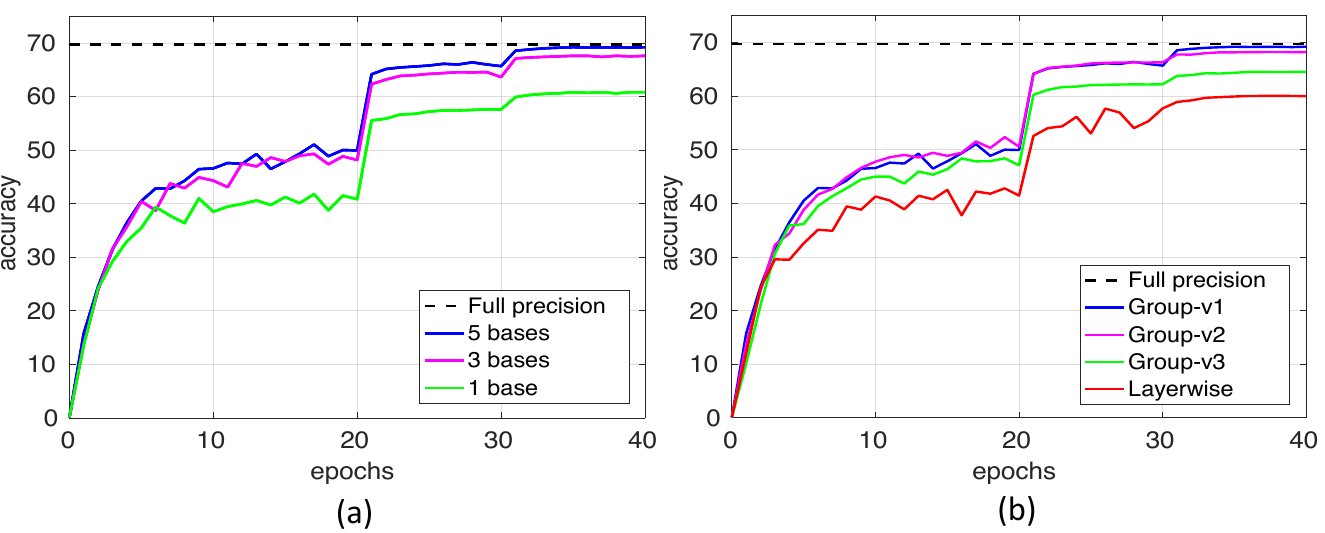}
		\end{tabular}
	}
	\caption{Top-1 validation accuracy of Group-Net with ResNet-18 on ImageNet. (a): The impact of the number of group bases. (b): The influence of different structure design.}
	\label{fig:convergence}
\end{figure*}
\vspace{-2mm}
We explore the difference between layer-wise and group-wise design strategies in Table~\ref{tab:bitwidth} and Fig.~\ref{fig:convergence} (b). By comparing the results, we find a 9.1\% significant performance increase between \emph{Group-Net v1} and \emph{Layerwise} under the same bitwidth. Note that the \emph{Layerwise} approach have similarities between tensor approximation methods in~\cite{guo2017network,  li2017performance, lin2017towards} and the differences are described in Sec.~\ref{sec:groupwise}. It strongly shows the necessity for employing the group-wise design strategy to get promising results. It also proves the importance by supressing the cumulative quantization error while making precise output tensor approximation. Moreover, we also speculate that this significant gain is partly due to the preserved block structure within the group bases. Interestingly, group-wise based approaches also converge more stably than the layer-wise one. We further provide the comparison on the network without residual connections (\ie, AlexNet) in Sec. S1 in the supplementary material.

\vspace{-3mm}
\subsection{Evaluation on CIFAR-100}
We also report our results with AlexNet on the CIFAR-100 dataset in Table~\ref{tab:cifar100}. We use the same group-wise strategy described in Sec.~\ref{sec:imagenet}. DOREFA-Net and EL-Net still use 2-bit weights and 2-bit activations. We can observe that out result outperforms all the competing approaches on the small dataset, which proves the robustness of the group-wise feature reconstruction strategy and the generalization ability of the proposed approach. 

\setlength{\abovedisplayskip}{1pt} 
\setlength{\belowdisplayskip}{1pt}
\begin{table*}[h]
	\centering
	\scalebox{0.9}
	{
		\begin{tabular}{c c c c c c}
			\hline
			Model &Full-precision&DOREFA &EL-Net&ABC-Net &Ours\\\hline
			\hline
			Top-1 & 65.9\% & 63.4\% &64.6\% &- &\bf{65.4}\%  \\
			Top-5  &88.1\% & 87.5\% &87.8\% &- &\bf{88.4}\% \\\hline
	\end{tabular}}
	\caption{Validation accuracy of Group-Net with AlexNet on CIFAR-100.}
	\label{tab:cifar100}
\end{table*}

\section{Conclusion}

In this paper, we have begun to explore highly efficient and accurate CNN architectures with binary weights and low-precision activations. Specifically, we have proposed to directly decompose the full-precision network into multiple groups and each group is approximated using a set of low-precision bases which can be optimized in an end-to-end manner. It is much more flexible than the previous approaches that operate layer-wise and can be integrated with the neural architecture search for exploring the optimal structure. The low-precision multi-branch group-wise structure can also be implemented in parallel and bring great benefits for accelerating test-time inference on specialized hardware. Experimental results have proved the robustness of the proposed approach on the ImageNet classification task. We further expect to generalize the work to other computer vision tasks.

\appendix

	\section{Layer-wise vs. group-wise on AlexNet}
	\begin{table*}[h]
		\centering
		\scalebox{0.9}
		{
			\begin{tabular}{c c c c}
				\hline
				Model& Full-precision &Layerwise &Group-S1   \\	
				\hline
				Top-1 & 57.2\%  &54.2\% &\bf{57.3}\%  \\
				Top-5  & 80.4 \%  &77.6\% &\bf{80.1}\% \\\hline	
		\end{tabular}}
		\caption{Layer-wise vs. group-wise with AlexNet on ImageNet.}
		\label{tab:compare}
	\end{table*}
	\begin{figure*}[h]
	\centering
	\resizebox{0.6\linewidth}{!}
	{
		\begin{tabular}{c}
			\includegraphics{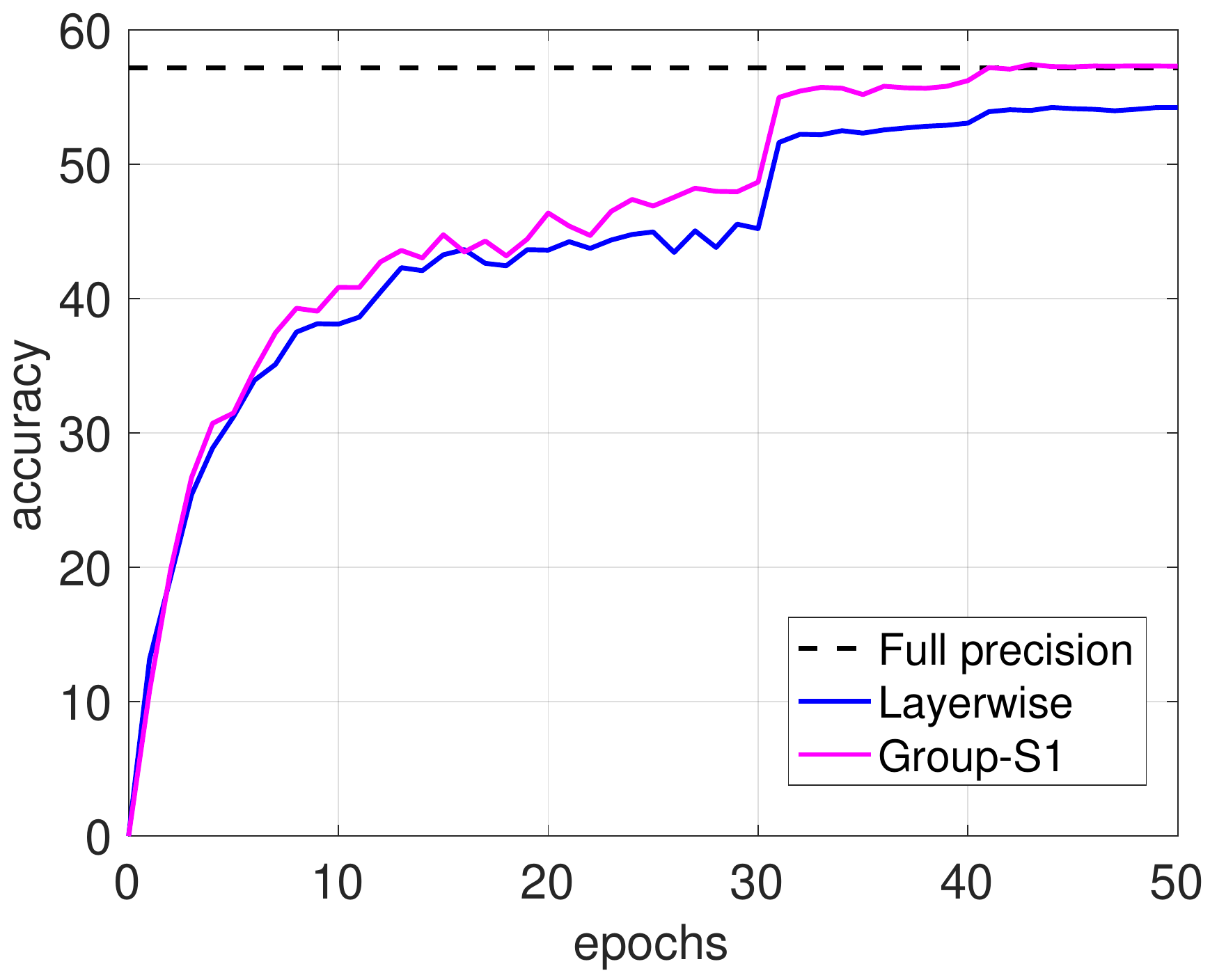}
		\end{tabular}
	}
	\caption{Top-1 validation accuracy of \emph{Layerwise} and \emph{Group-S1} with AlexNet on ImageNet.}
	\label{fig:alexnet}
\end{figure*}
	In this section, we empirically analyze the difference between layer-wise and group-wise design strategies on the plain network AlexNet. We perform experiments on the ImageNet dataset. As described in Sec.~4.2 in the paper, we treat two subsequent convolutional or dense layers as a group which we call \emph{Group-S1}. We also provide the results of \emph{Layerwise} approach described in Sec.~4 in the paper.
	From Table~\ref{tab:compare}, we can observe that the group-wise quantization strategy outperforms the \emph{Layerwise} approach with significant margins. It proves that the group-wise feature approximation approach is not only effective for residual architectures but also works well on plain network architectures without skip connections. We also illustrate the convergence curves for these two approaches in Fig.~\ref{fig:alexnet}.

	\newpage

      \section{Summary of the training algorithm}

		\begin{algorithm}[]
		\KwIn{A minibatch of inputs and targets, current $p$-th group full-precision weights $\{ {\mathbb{W}}_1^p,{\mathbb{W}}_2^p,...,{\mathbb{W}}_M^p\}$ where $\mathbb{W}_m^p$ represents the $m$-th group base consists of weights $\{ {\bf{w}}_{m1}^p, {\bf{w}}_{m2}^p,...,{\bf{w}}_{m{N_p}}^p\}$ for $N_p$ layers within the base. The initial learning rate is $ \eta$ and the decay factor $\varepsilon$. From the definition, the number of layers $L = \sum\limits_{p = 1}^P {{N_p}}$.}
		\KwOut{Parameters of all groups.}
        \textbf{Forward Propagation} \\
        \For {$p$ = $1$ to $P$}
        {   \For {$i$ = $1$ to $N_p$}
        	{    
        	      Compute binary weight ${\bf{b}}_{mi}^p$ using weight quantization function in Eq. 1  with ${{\bf{w}}_{mi}^p}$, $m = 1,...,M$; \\ 
                  ${\bf{x}}_{mi}^p \leftarrow {\rm{Conv}}({\bx}_{m(i - 1)}^p,{\bf{b}}_{mi}^p)$; \\
        	      Apply batch normalization; \\ 
        	      \If{$i<N_p$}
        	      {
        	      Apply activation quantization function in Eq. 2 to ${\bf{x}}_{mi}^p$ to obtain $ {\bf{\widetilde{x}}}_{mi}^p$;  
        	     }
                 \Else {
                 	Obtain the $p$-th group's output by the aggregating operation ${\bf{s}}^p= \sum\limits_{m = 1}^M {{\theta _m^{p}}{\bf{x}}_{m{N_p}}^p}$; \\
                 	 Quantize the activations ${\bf{s}}^p$ to $ {\bf{\widetilde{s}}}^p$ using Eq. 2; \\
                 	 Obtain the $(p+1)$-th group's input ${\bx}_{m0}^{p + 1} \leftarrow {\bs^p}$;
                          }
                   Optionally apply pooling;
        	  }     

        }
        Obtain the $P$-th group's output ${\bx_L} \leftarrow {\bs^P}$; \\
        Compute the loss $\ell$ using ${\bx_L}$ and the targets;  \\
        \textbf{Back Propagation} \\     
            Compute gradients of the final layer ${g_{{\bx_L}}} = \frac{{\partial \ell }}{{\partial {\bx_L}}}$; \\
            \For {$p$ = $P$ to $1$}
            {
                \For {$i$ = $N_p$ to $1$}
                 {Compute ${g_{\bx_{m(i - 1)}^p}}$ using ${g_{\bx_{mi}^p}}$ and ${\bf{b}}_{mi}^p$; \\
                   Compute ${g_{{\bf{b}}_{mi}^p}}$ using ${g_{{\bx}_{mi}^p}}$ and $\bx_{m(i - 1)}^p$; \\
                 }
                  \If{$p>1$}
                  {Compute ${g_{\theta _m^{p - 1}}}$ using ${\bf{x}}_{m{N_{p - 1}}}^{p - 1}$ and ${{g_{{{\bf{s}}^{p-1}}}}}$, where ${g_{{{\bf{s}}^{p-1}}}} \approx {g_{{\bs^{p-1}}}}$ according to Eq. 2;}
             }
         \textbf{Update parameters using SGD} \\
           \For{$p$ = $1$ to $P$} 
           {    \For {$i$ = $1$ to $N_p$}
           	      {
           	         Compute ${g_{{\bf{w}}_{mi}^p}}$ with ${g_{{\bf{b}}_{mi}^p}}$ using Eq. 1, $m=1, 2, ..., M$; \\
           	         Update  ${\bf{w}}_{mi}^p$  with ${g_{{\bf{w}}_{mi}^p}}$ using SGD, $m=1, 2, ..., M$; \\  
           	    }        
               Update $\theta _m^p$ with ${g_{\theta _m^{p}}}$ using SGD, $m=1, 2, ..., M$; \\     
       }
       $\eta  \leftarrow \varepsilon \eta$; \\
           
		\caption{Training a $L$ layers feedforward neural network. We partition it into $P$ groups and each group has $M$ bases. $k$ is the bitwidth of activations. Weights are all binary. Note that when $k>1$, $\rm{Conv()}$ is the fixed-point accumulations. When $k=1$, as described in Sec. 3.1, $\rm{Conv()}$ is the XNOR bit operations and we use the quantization approach in XNOR-Net [29].} 
		\label{algo:construct_blocks}
	\end{algorithm}

\clearpage

\bibliographystyle{abbrv}
{
	\small
	\bibliography{reference}
}

\end{document}